\documentclass[10pt,twocolumn,letterpaper]{article}

\usepackage{cvpr}
\usepackage{times}
\usepackage{epsfig}
\usepackage{graphicx}
\usepackage{amsmath}
\usepackage{amssymb}
\usepackage{multirow}

\usepackage[pagebackref=true,breaklinks=true,letterpaper=true,colorlinks,bookmarks=false]{hyperref}

\cvprfinalcopy 

\def\cvprPaperID{} 

\begin{document}

\title{Transferable Interactiveness Knowledge for \\Human-Object Interaction Detection}
\author{Yong-Lu Li,~Siyuan Zhou,~Xijie Huang,~Liang Xu,~Ze Ma,~Hao-Shu Fang,~Yan-Feng Wang,~Cewu Lu\thanks{Cewu Lu is the corresponding author, he is also a member of Department of Computer Science and Engineering, Shanghai Jiao Tong University, MoE Key Lab of Artificial Intelligence, AI Institute, Shanghai Jiao Tong University, and SJTU-SenseTime AI lab.}\\
Shanghai Jiao Tong University\\
{\tt\small \{yonglu\_li, ssluvble, otaku\_huang, liangxu, maze1234556\}@sjtu.edu.cn }\\
{\tt\small fhaoshu@gmail.com,}
{\tt\small wangyanfeng@sjtu.edu.cn,}
{\tt\small lucewu@sjtu.edu.cn}
}

\maketitle

\begin{abstract}
Human-Object Interaction (HOI) Detection is an important problem to understand how humans interact with objects. In this paper, we explore \textbf{Interactiveness Knowledge} which indicates whether human and object interact with each other or not. We found that interactiveness knowledge can be learned across HOI datasets, regardless of HOI category settings. Our core idea is to exploit an Interactiveness Network to learn the general interactiveness knowledge from multiple HOI datasets and perform Non-Interaction Suppression before HOI classification in inference. On account of the generalization of interactiveness, interactiveness network is a transferable knowledge learner and can be cooperated with any HOI detection models to achieve desirable results.
We extensively evaluate the proposed method on HICO-DET and V-COCO datasets. Our framework outperforms state-of-the-art HOI detection results by a great margin, verifying its efficacy and flexibility. Code is available at \url{https://github.com/DirtyHarryLYL/Transferable-Interactiveness-Network}.
\end{abstract}
   
\section{Introduction}
Human-Object Interaction (HOI) detection retrieves human and object locations and infers the interaction classes from still image. As a sub-task of visual relationship~\cite{visualgenome,Lu2016Visual}, HOI is strongly related to the human body and object understanding~\cite{fang2018learning,xiu2018pose,fang2018weakly,faster-rcnn,lu2018beyond,maskrcnn,xu2018srda}. It is crucial for behavior understanding and can facilitate activity understanding~\cite{activitynet,deeprnn}, imitation learning~\cite{immitation}, etc. Recently, impressive progress has been made by utilizing Deep Neural Networks (DNNs) in this area~\cite{hicodet,Gkioxari2017Detecting,qi2018learning,gao2018ican}.
\begin{figure}[h]
	\begin{center}
		\includegraphics[width=0.48\textwidth]{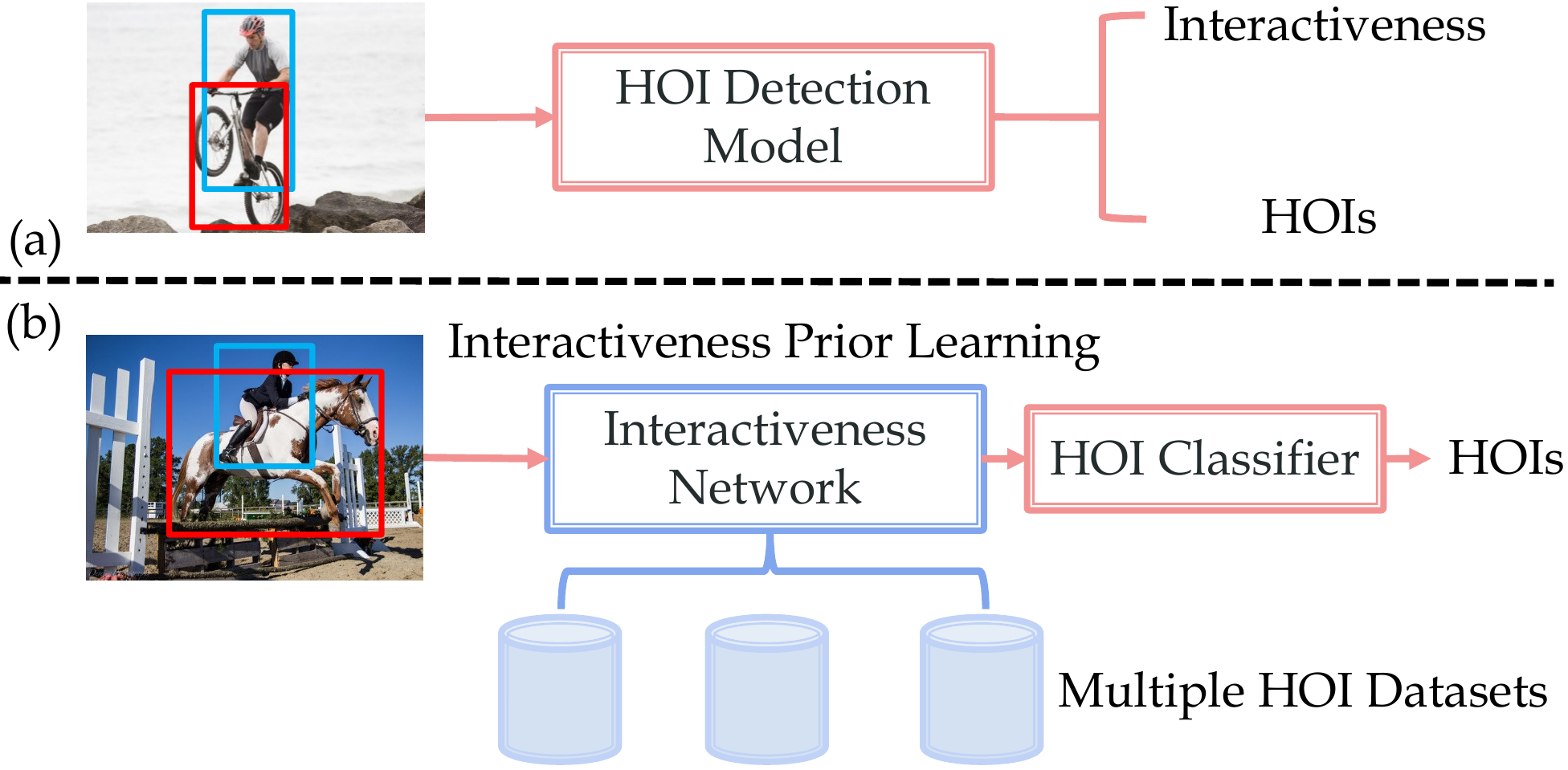}
	\end{center}
	\caption{Interactiveness Knowledge Learning. (a) HOI datasets contain implicit interactiveness knowledge. We can learn it better by performing explicit interactiveness discrimination, and utilize it to improve the HOI detection performance. (b) Interactiveness knowledge is beyond the HOI categories and can be learned across datasets, which can bring greater performance improvement.}
	\label{Figure:pipeline}
	\vspace{-0.3cm}
\end{figure}

Generally, human and objects need to be detected first. Given an image and its detections, human and objects are often paired exhaustively~\cite{Gkioxari2017Detecting,gao2018ican,qi2018learning}. HOI detection task aims to classify these pairs as different HOI categories. Previous one-stage methods~\cite{hicodet,Gkioxari2017Detecting,gao2018ican,vcoco,qi2018learning} directly classify a pair as specific HOIs. These methods actually predict \emph{interactiveness} implicitly at the same time, where interactiveness indicates whether a human-object pair is interactive. For example, when a pair is classified as HOI ``eat apple'', we can implicitly predict that it is interactive. 

Though interactiveness is an essential element for HOI detection, we neglected to study how to utilize it and improve its learning. In comparison to HOI categories, interactiveness conveys more basic information. 
Such attribute makes it easier for interactiveness to transfer across datasets. 
Based on this inspiration, we propose a \textbf{Interactiveness Knowledge} learning method as seen in Figure \ref{Figure:pipeline}. 
With our framework, interactiveness can be learned across datasets and applied to any specific dataset.
By utilizing interactiveness, we take two stages to identify HOIs: we first discriminate a human-object pair as interactive or not and then classify it as specific HOIs. 
Compared to previous one-stage method~\cite{hicodet,Gkioxari2017Detecting,gao2018ican,vcoco,qi2018learning}, we take advantage of powerful interactiveness knowledge that incorporates more information from other datasets. Thus our method can decrease the false positives significantly. Additionally, after the interactiveness filtering in the first stage, we do not need to handle a large number of non-interactive pairs which are overwhelmingly more than interactive ones.

\begin{figure}[!ht]
	\begin{center}
		\includegraphics[width=0.45\textwidth]{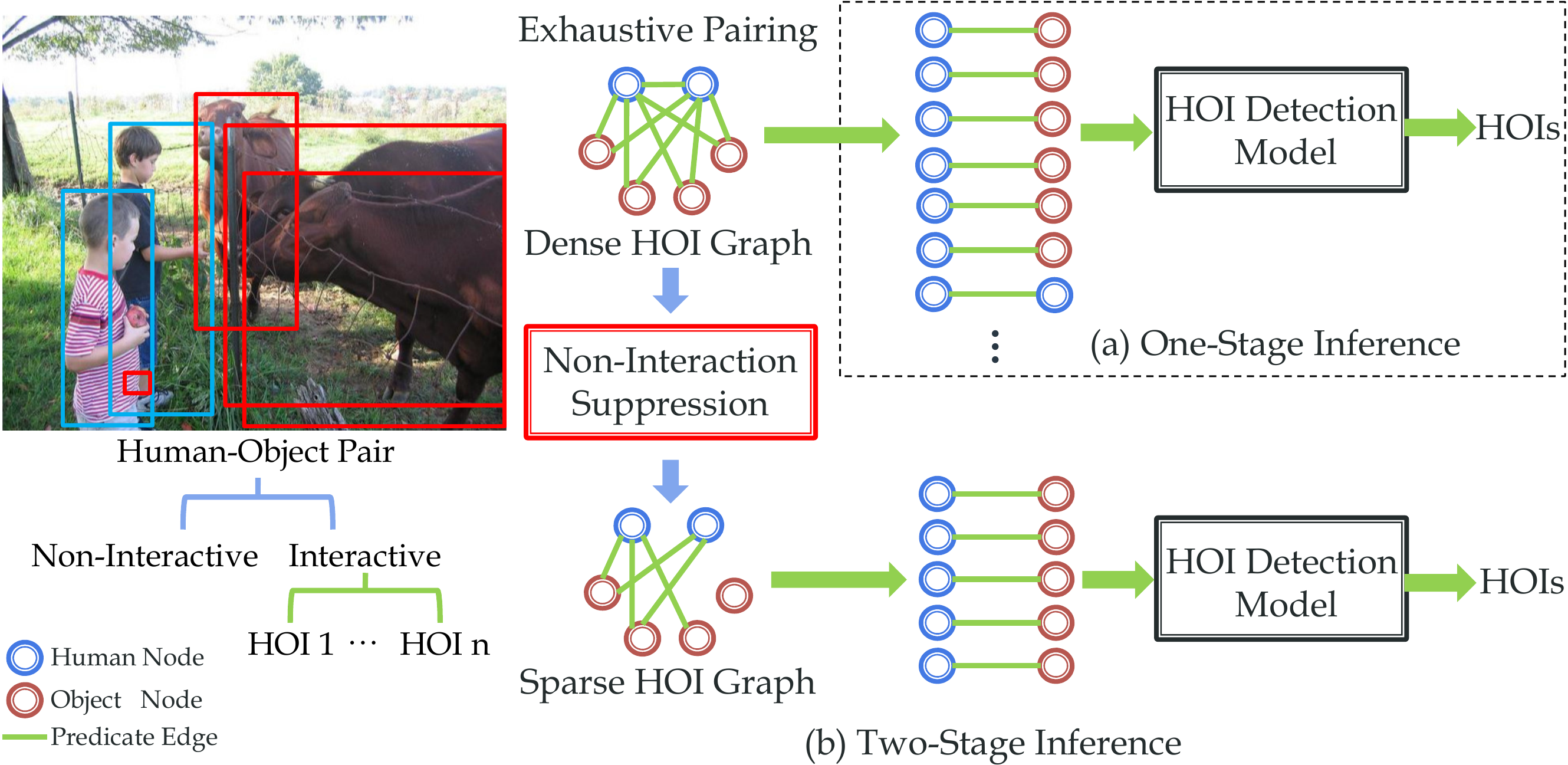}
	\end{center}
	\caption{HOIs within an image can be represented as a HOI graph. Human and object can be seen as nodes, whilst the interactions are represented as edges. Exhaustive pairing of all nodes would import overmuch non-interactive edges and do damage to detection performance. Our Non-Interaction Suppression can effectively reduce non-interactive pairs. Thus the dense graph would be converted to a sparse graph and then be classified.}
	\label{Figure:graph-refine}
	\vspace{-0.3cm}
\end{figure}

In this paper, we proposed a novel two-stage method to classify pairs hierarchically as shown in Figure \ref{Figure:graph-refine}. We introduce an interactiveness network which can be combined with any HOI detection model. We set a hierarchical logical strategy: by utilizing binary interactiveness labels, interactiveness network will bring in a strong supervised constraint which refines the framework in training and learns the interactiveness from multiple datasets. In testing, interactiveness network performs Non-Interaction Suppression (NIS) first. Then the HOI detection model will classify the remaining pairs as specific HOIs, where non-interactive pairs have been decreased significantly. Moreover, if the model classifies a pair as specific HOIs, it should figure out that the pair is interactive simultaneously. Such two-stage prediction will alleviate the learning difficulty and bring in hierarchical predictions. For special attention, interactiveness offers extra information to help HOI classification and is independent of HOI category settings. That means it can be transferred across datasets and utilized to enhance HOI models designed for different HOI settings.

We perform extensive experiments on HICO-DET~\cite{hicodet}, V-COCO~\cite{vcoco} datasets. Our method cooperated with transferred interactiveness outperforms the state-of-the-art methods by \textbf{2.38}, \textbf{3.06}, and \textbf{2.17} mAP on three Default category sets on HICO-DET, \textbf{4.0} and \textbf{3.4} mAP on V-COCO.

\section{Related Works}

\noindent{\bf Visual Relationship Detection.} Visual relationship detection~\cite{Sadeghi2012Recognition,Lu2016Visual,visualgenome,Yatskar2016Situation} aims to detect the objects and classify their relationships simultaneously. In ~\cite{Lu2016Visual}, Lu \etal proposed a relationship dataset VRD and an approach combined with language priors. Predicates within relationship triplet $\langle subject, predicate, object \rangle$ include actions, verbs, spatial and preposition vocabularies. Such vocabulary setting and severe long-tail issue within the dataset make this task quite difficult. Large-scale dataset Visual Genome~\cite{visualgenome} is then proposed to promote studies in this problem. Recent works ~\cite{xu2017scene,vtranse,yin2018zoom,yang2018graph} put attention on more effective and efficient visual feature extraction and try to exploit semantic information to refine the relationship detection.
\begin{figure*}[!ht]
	\begin{center}
		\includegraphics[width=0.8\textwidth]{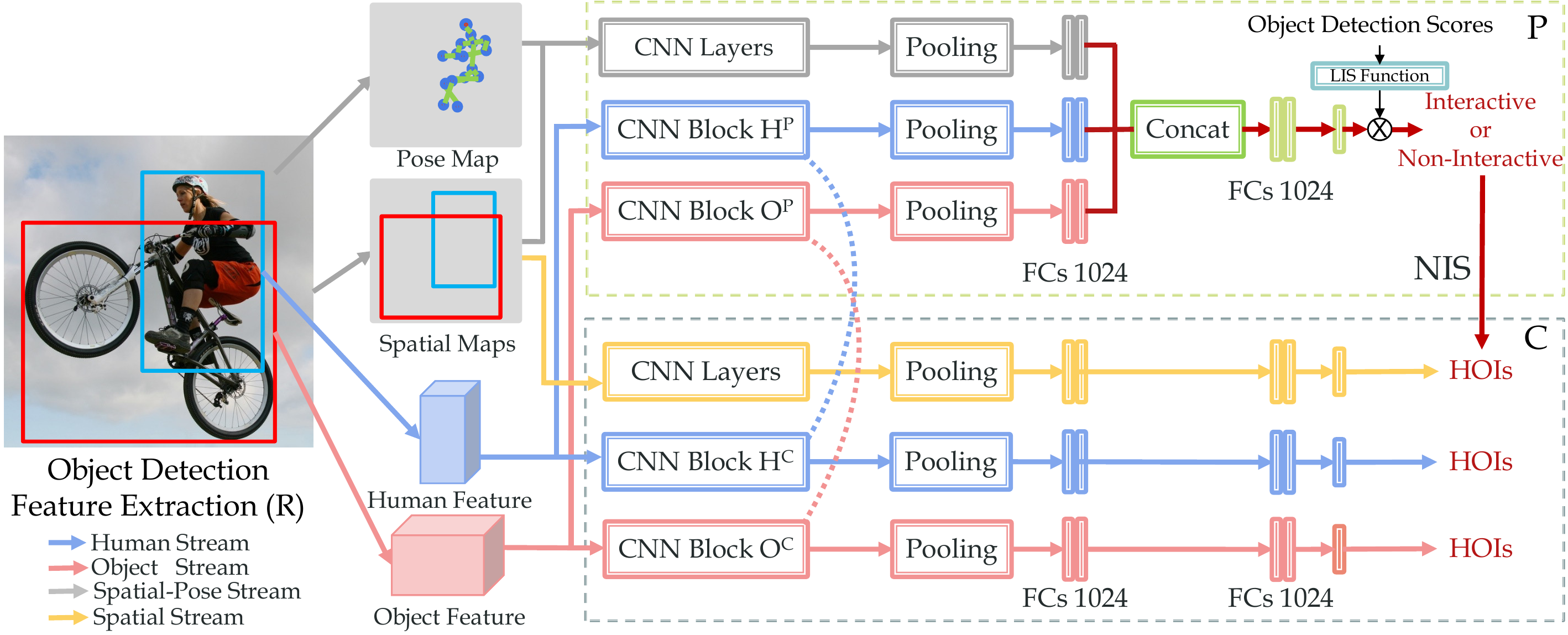}
	\end{center}
	\caption{Overview of our framework. Interactiveness network $\mathbf{P}$ can cooperate with any HOI models (referred as $\mathbf{C}$). $\mathbf{P}$ employs human, object and spatial-pose streams to extract features from human and object appearance, spatial locations and human pose information. The outputs of three streams are concatenated and inputted to the interactiveness discriminator. 
	When cooperated with multi-stream $\mathbf{C}$ such as~\cite{hicodet,gao2018ican} (human, object, and spatial streams),
	$H^{\mathbf{P}}$ and $O^{\mathbf{P}}$ in $\mathbf{P}$ can share weights (dotted lines) with $H^{\mathbf{C}}$ and $O^{\mathbf{C}}$ in $\mathbf{C}$ during joint training. In this work, these four blocks are all residual blocks~\cite{resnet}.
	LIS and NIS will be detailed in Section \ref{sec:d} and Section \ref{sec:test}.}
	\label{Figure:overview}
\end{figure*}

\noindent{\bf Human-Object Interaction Detection.} Human-Object Interaction~\cite{Wang2006Unsupervised,Yang2010Recognizing,Ikizler2008Recognizing} is essential to understand human-centric interaction with objects. Recently several large-scale datasets, such as V-COCO~\cite{vcoco}, HICO-DET~\cite{hicodet}, HCVRD~\cite{hcvrd}, were proposed for the exploration of HOI detection.
Different from HOI recognition~\cite{Fang2018Pairwise,Delaitre2010Recognizing,hico,Chao2014Predicting,Mallya2016Learning} which is an image level classification problem, HOI detection needs to detect interactive human-object pairs and classify their interactions at instance level. 
With the assistance of DNNs and large-scale datasets, recently methods have made significant progress.
Chao \etal~\cite{hicodet} proposed a multi-stream model combining visual features, spatial locations to help tackle this problem.
To address the long tail issue, Shen \etal~\cite{Shen2018Scaling} studied zero-shot learning problem and predicted the verb and object separately.
In~\cite{Gkioxari2017Detecting}, an action specific density map estimation method is introduced to locate objects interacted with human. 
In~\cite{qi2018learning}, Qi \etal proposed GPNN incorporating DNN and graphical model, which uses message parsing to iteratively update states and classifies all possible pairs/edges.
Gao \etal~\cite{gao2018ican} exploited an instance centric attention module to enhance the information from the interest region and facilitate the HOI classification.
Generally, these methods inference in one-stage and may suffer from severe non-interactive pair domination problem. To address this issue, we utilize interactiveness to explicitly discriminate non-interactive pairs and suppress them before HOI classification.
\vspace{-0.3cm}

\section{Preliminary}
HOI representation can be described as a graph model~\cite{qi2018learning, xu2017scene} as seen in Figure \ref{Figure:graph-refine}. Instances and relations are expressed as nodes and edges respectively. With exhaustive pairing~\cite{Gkioxari2017Detecting,gao2018ican}, HOI graph $\mathcal{G} = (\mathcal{V}, \mathcal{E})$ is dense connected, where $\mathcal{V}$ includes human node $\mathcal{V}_h$ and object node $\mathcal{V}_o$. 
Let $v_h \in \mathcal{V}_h$ and $v_o \in \mathcal{V}_o$ denote the human and object nodes. Thus edges $e \in \mathcal{E}$ are expressed as $e = (v_h, v_o) \in \mathcal{V}_h \times \mathcal{V}_o$. With $n$ nodes, exhaustive paring will generate a mass of edges. We aim to assign HOI (including no HOI) labels on those edges. Considering that a vast majority of non-interactive edges existing in $\mathcal{E}$ should be discarded, our goal is to seek a sparse $\mathcal{G}^{*}$ with corrected HOI labeling on its edges.

\section{Our Method}
\subsection{Overview}
As aforementioned, we introduce \textbf{Interactiveness Knowledge} to advance HOI detection performance. That is, explicitly discriminate the non-interactive pairs and suppress them before HOI classification. From the semantic point of view, interactiveness provides more general information than conventional HOI categories. Since any human-object pair can be assigned binary interactiveness labels according to the HOI annotations, \ie ``interactive'' or ``non-interactive'', interactiveness knowledge can be learned from multiple datasets with different HOI category settings and transferred to any specific datasets.

To exploit this cue, we proposed interactiveness network (interactiveness predictor, referred as $\mathbf{P}$) which utilizes interactiveness to reduce false positives caused by overmuch non-interactive pair candidates. 
Some conventional modules are also included, namely, Representation Network $\mathbf{R}$ (feature extractor) and Classification Network $\mathbf{C}$ (HOI classifier). 
$\mathbf{R}$ is responsible for feature extraction from detected instances. $\mathbf{C}$ utilizes node and edge features to perform HOI classification. Figure \ref{Figure:overview} is an overview of our framework which follows the hierarchical classification paradigm. Specifically, we first train $\mathbf{P}$ and $\mathbf{C}$ jointly to learn the interactiveness and HOIs knowledge. Under usual circumstances, the ratio of non-interactive edges is dominant within inputs. Hence $\mathbf{P}$ will bring a strong supervised signal to refine the framework. In testing, $\mathbf{P}$ is utilized in two stages. First, $\mathbf{P}$ evaluates the interactiveness of edges by exploiting the learned interactiveness knowledge, so we can convert the dense HOI graph to a sparse one. Second, combined with interactiveness score from $\mathbf{P}$, $\mathbf{C}$ will process the sparse graph and classify the remaining edges. 

In addition, on account of the generalization ability of interactiveness knowledge, it can be transferred with $\mathbf{P}$ across datasets (Section~\ref{sec:train-test}). Details of the framework architecture are illustrated in Section \ref{sec:ec} and \ref{sec:d}. The process of training and testing will be detailed in Section \ref{sec:train-test}.

\subsection{Representation and Classification Networks}
\label{sec:ec}
\noindent{\bf Human and Object Detection.} In HOI detection, human and object need to be detected first. In this work, we follow the setting of~\cite{gao2018ican} and employ the Detectron~\cite{detectron} with ResNet-50-FPN~\cite{fpn} to prepare bounding boxes and detection scores. Before post-processing, detection results will be filtered by the detection score thresholds first.

\noindent{\bf Representation Network.} In previous methods~\cite{hicodet,Gkioxari2017Detecting,gao2018ican}, $\mathbf{R}$ is often modified from object detector such as Fast R-CNN~\cite{fastrcnn} or Faster R-CNN~\cite{faster-rcnn}. We also exploited a Faster R-CNN~\cite{faster-rcnn} with ResNet-50~\cite{resnet} based $\mathbf{R}$ here. During training and testing, $\mathbf{R}$ is frozen and acts as a feature extractor.  
Given the detected bounding boxes, we produce human and object features by cropping ROI pooling feature maps according to box coordinates. 
\begin{figure}[!ht]
	\begin{center}
		\includegraphics[width=0.48\textwidth]{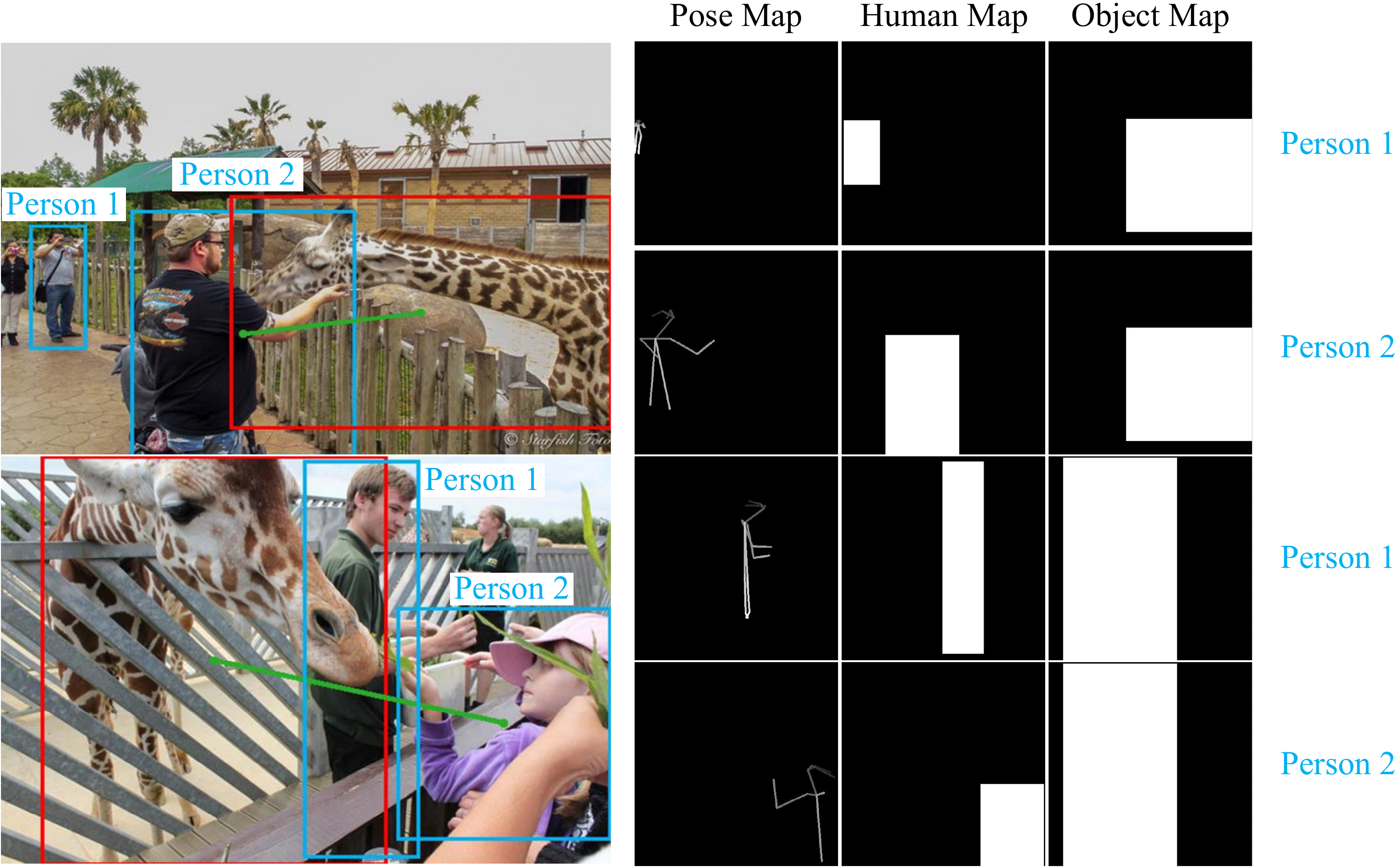}
	\end{center}
	\caption{Inputs of the spatial-pose stream. Three kinds of maps are included: pose map, human map and object map. Person 2 in two images both have interaction ``feed'' with giraffes. But two pairs of Person 1 and giraffe are all non-interactive. Their poses and locations are helpful for the interactiveness discrimination.}
	\label{Figure:d work reason}
	\vspace{-0.3cm}
\end{figure}

\noindent{\bf HOI Classification Network.} As for $\mathbf{C}$, multi-stream architecture and late fusion strategy are frequently used and approved effective~\cite{hicodet,gao2018ican}. 
Follow~\cite{hicodet,gao2018ican}, for our classification network $\mathbf{C}$, we utilize a human stream and an object stream to extract human, object and context features. Within each stream, a residual block~\cite{resnet} (denoted as $H^{C}$, $O^{C}$, seen in Figure~\ref{Figure:overview}) with pooling layer and fully connected layers (FCs) are adopted.
Moreover, an extra spatial stream~\cite{hicodet} is adopted to encode the spatial locations of instances. Its input is a two-channel tensor consisting of a human map and an object map, shown in Figure \ref{Figure:d work reason}. Human and object maps are all 64x64 and obtained from the human-object union box. In the human channel, the value is 1 in the human bounding box and 0 in other areas. The object channel is similar which has value 1 in the object bounding box and 0 elsewhere.
Following the late fusion strategy, each stream will first perform HOI classification, then three prediction scores will be fused by element-wise sum in the same proportion to produce the final result of $\mathbf{C}$.

\subsection{Interactiveness Network}
\label{sec:d}
Interactiveness needs to be learned by extracting and combining essential information. The visual appearance of human and object are obviously required.
Besides, interactive and non-interactive pairs also have other distinguishing features, e.g. spatial location and human pose information.
For example, in the upper image of Figure \ref{Figure:d work reason}, Person 1 and the giraffe far from him are not interactive. Their spatial maps~\cite{hicodet} can provide pieces of evidence to help with classification. 
Furthermore, pose information is also helpful. In the lower image, although two people are both close to the giraffe, only Person 2 and the giraffe are interactive. The arm of Person 2 is uplift and touching the giraffe. Whilst Person 1 is back on to the giraffe, and his pose is quite different from the typical pose of ``feed''.

\begin{figure}[!ht]
	\begin{center}
		\includegraphics[width=0.45\textwidth]{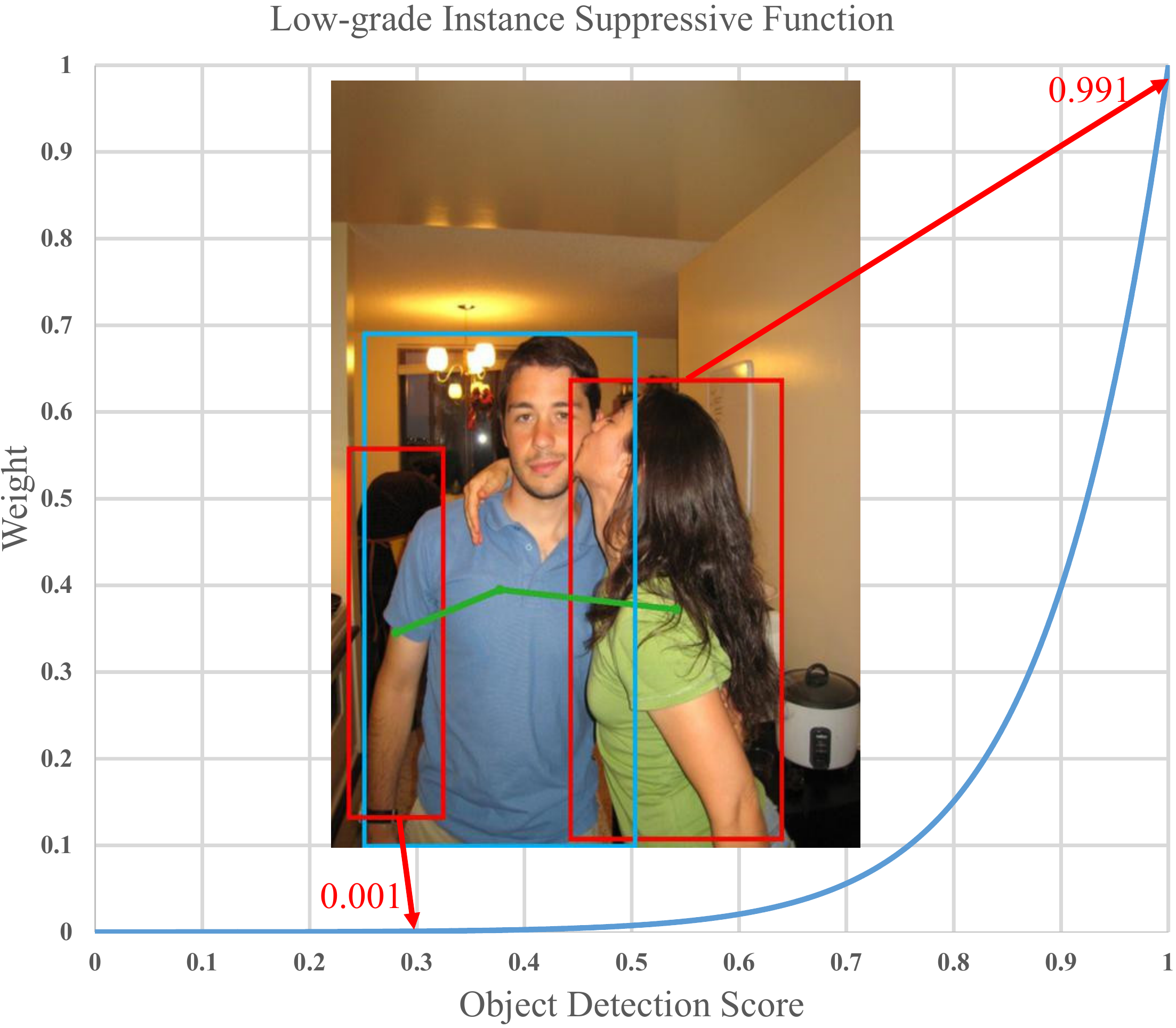}
	\end{center}
	\caption{The illustration of $\mathcal{P}(\cdot)$ within Low-grade Suppressive Function. Its input is object detection score. High-grade detected objects will be emphasized and distinguished with low-grade ones. In addition, $\mathcal{P}(0)=5.15E-05$ and $\mathcal{P}(1)=9.99E-01$. }
	\label{Figure:logictic}
	\vspace{-0.3cm}
\end{figure}
Based on these reasons, the combination of visual appearance, spatial location and human pose information is key to interactiveness discrimination. Hence $\mathbf{P}$ needs to encode these key elements together to learn the interactiveness knowledge. A natural choice is the multi-stream architecture as presented: human, object and spatial-pose streams.

\noindent{\bf Human and Object stream.} For human and object appearance, we extract ROI pooling features from representation network $\mathbf{R}$, then input them into residual blocks $H^{\mathbf{P}}$ and $O^{\mathbf{P}}$, respectively. The architecture of $H^{\mathbf{P}}$ and $O^{\mathbf{P}}$ are same as $H^{\mathbf{C}}$ and $O^{\mathbf{C}}$ (Figure~\ref{Figure:overview}). Through subsequent global average pooling and FCs, the output features of two streams are denoted as $f_h$ and $f_o$, respectively. 

\noindent{\bf Spatial-Pose Stream.} Different from~\cite{hicodet}, our spatial-pose stream input includes a special 64x64 pose map.
Given the union box of each human and his/her paired object, we employ pose estimation~\cite{fang2017rmpe,crowdpose} to estimate his/her 17 keypoints (in COCO format~\cite{coco}). Then, we link the keypoints with lines of different gray value ranging from 0.15 to 0.95 to represent different body parts, which implicitly encodes the pose features. Whilst the other area is set as 0.
Finally, we reshape the union box to 64x64 to construct the pose map. 
We concatenate the pose map with human and object maps which are the same as those in the spatial stream of $\mathbf{C}$. This forms the input for our spatial-pose stream.
Next, we exploit two convolutional layers with max pooling and two 1024 sized FCs to extract the feature $f_{sp}$ of three maps. Last, the output will be concatenated with the outputs of human and object streams for interactiveness discrimination.

Given a HOI graph $\mathcal{G}$ with all possible edges, $\mathbf{P}$ will evaluate the interactiveness of pair $(v_h, v_o)$ based on learned knowledge, and gives confidence:
\begin{eqnarray}
    s^{P}_{(h,o)} = f_{\mathbf{P}}(f_h,f_o,f_{sp}) \ast L(s_h, s_o),
\label{eq:d_score}
\end{eqnarray}
where  $L(s_h, s_o)$  is a novel weight function named Low-grade Instance Suppressive Function (LIS). It takes the human and object detection scores $s_h,s_o$ as inputs:
\begin{eqnarray}
    L(s_h, s_o) = \mathcal{P}(s_h) \ast \mathcal{P}(s_o),
\label{eq:lis}
\end{eqnarray}
where
\begin{eqnarray}
    \mathcal{P}(x) = \frac{T}{1+e^{(k-wx)}},
\label{eq:D_sp_func}
\end{eqnarray}
$\mathcal{P}(\cdot)$ is a part of the logistic function, the value of $\mathbf{T}$, $\mathbf{k}$ and $\mathbf{w}$ will be determined by data-driven manner. Figure \ref{Figure:logictic} depicts the curve of $\mathcal{P}(\cdot)$ whose domain definition is $(0,1)$. Bounding boxes will have low weight till their score is higher than a threshold. Previous works~\cite{gao2018ican,Gkioxari2017Detecting} often directly multiply detection scores by the final classification score. But they cannot notably emphasize the differentiation between high quality and inaccurate detection results. LIS has the ability to enhance the differentiation between high and low grade object detections as shown in Figure \ref{Figure:logictic}. 

\noindent{\bf Weights Sharing Strategy.} An additional benefit of our interactiveness network is that, if cooperated with multi-stream HOI detection model $\mathbf{C}$, $\mathbf{P}$ can share the weights of convolutional blocks with the ones in $\mathbf{C}$. As shown in Figure \ref{Figure:overview}, blocks $H^{\mathbf{P}}$ and $O^{\mathbf{P}}$ can share weights with $H^C$ and $O^C$ in the joint training. This weights sharing strategy can guarantee information sharing and better optimization of $\mathbf{P}$ and $\mathbf{C}$ in the multi-task training. 

\subsection{Interactiveness Knowledge Transfer Training}
\label{sec:train-test}
With $\mathbf{R}$, $\mathbf{P}$ and $\mathbf{C}$, our framework has two modes of utilization: hierarchical joint training in \emph{Default Mode}, and interactiveness transfer training in \emph{Transfer Learning Mode}.

\begin{figure}[!ht]
	\begin{center}
		\includegraphics[width=0.48\textwidth]{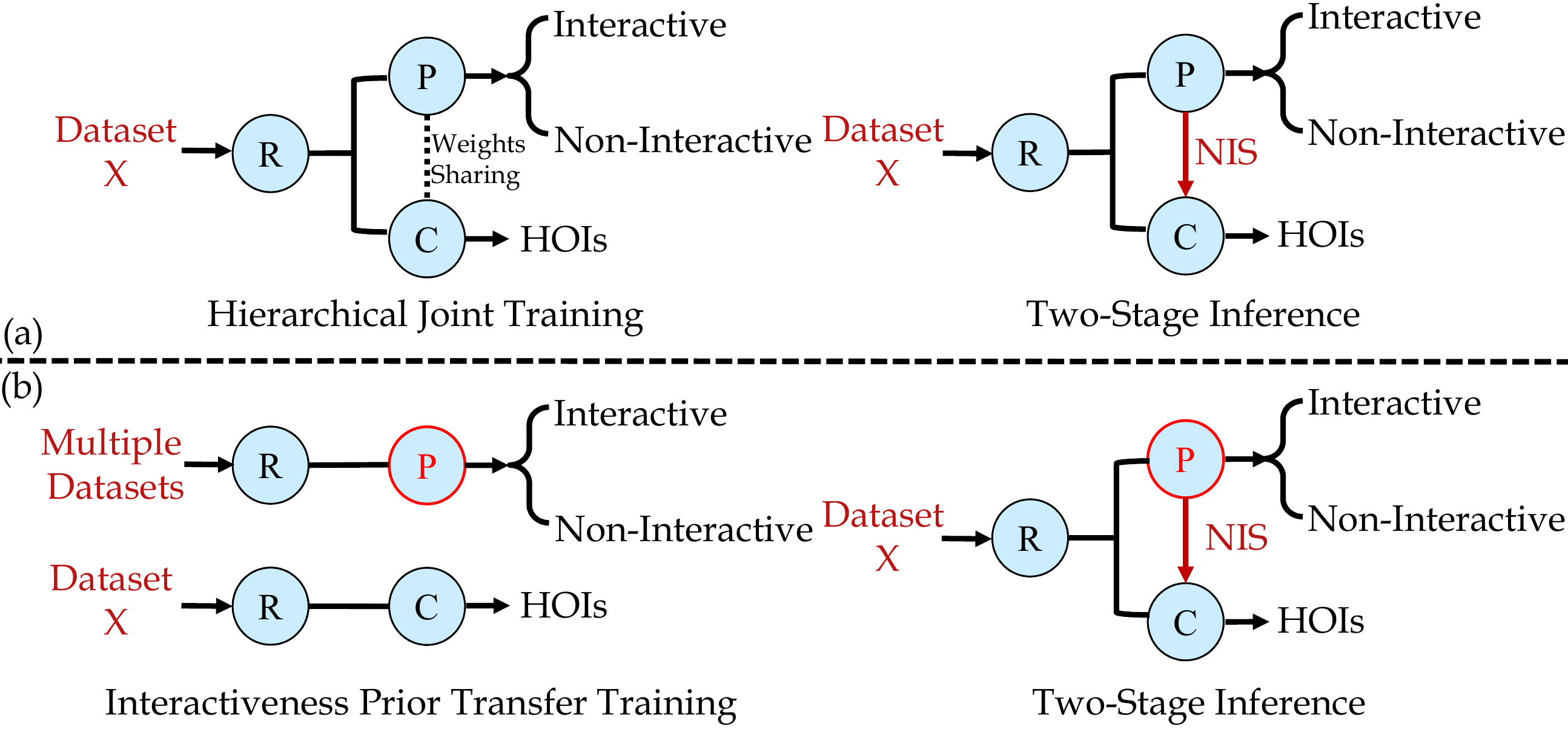}
	\end{center}
	\caption{The schemes for training and testing. 
	(a) In Default Mode, $\mathbf{P}$ and $\mathbf{C}$ are first trained jointly with weights sharing on the same dataset.
	(b) In Transfer Learning Mode, $\mathbf{P}$ can learn interactiveness knowledge across datasets and cooperates with multiple $\mathbf{C}$s trained on different datasets. 
	In testing, our framework infers in two stages, \ie $\mathbf{P}$ performs interactiveness discrimination at first, then $\mathbf{C}$ classifies the remaining edges/pairs.}
	\label{Figure:train-test}
	\vspace{-0.3cm}
\end{figure}

\noindent{\bf Hierarchical Joint Training.} In \emph{Default Mode}, we introduce our hierarchical joint training scheme, as illustrated in Figure \ref{Figure:train-test} (a). By adding a supervisor $\mathbf{P}$, our framework works in an unconventional training mode. 
To be specific, the framework is trained with hierarchical classification tasks, \ie explicit interactiveness discrimination and HOI classification.
The objective function of the framework can be expressed as:
\begin{eqnarray}
    \mathcal{L} = \mathcal{L}^{\mathbf{C}} + \mathcal{L}^{\mathbf{P}},
\label{eq:loss}
\end{eqnarray}
where $\mathcal{L}^{\mathbf{C}}$ denotes the HOI classification cross entropy loss, while $\mathcal{L}^{\mathbf{P}}$ is the binary classification cross entropy loss.

Different from one-stage methods, additional interactiveness discrimination enforces the model to learn interactiveness knowledge, which can bring more powerful supervised constraints. Namely, when a pair is predicted as specific HOIs such as ``cut cake'', $\mathbf{P}$ must give the prediction ``interactive'' simultaneously. 
Experiment results (Section \ref{sec:ablation}) prove that interactiveness knowledge learning can effectively refine the training and improve the performance. 
The framework in Default Mode is called ``$\mathbf{R}\mathbf{P}_{D}\mathbf{C}_{D}$'' in the following, where ``\emph{D}'' indicates ``Default''.

\noindent{\bf Interactiveness Knowledge Transfer Training.} Noting that $\mathbf{P}$ only needs binary labels which are beyond the HOI classes, so interactiveness is transferable and reusable. In \emph{Transfer Learning Mode}, $\mathbf{P}$ can be used as a transferable knowledge learner to learn interactiveness from multiple datasets and be applied to each of them respectively, as illustrated in Figure \ref{Figure:train-test} (b). On the contrary, $\mathbf{C}$ must be trained on a single dataset once a time considering the variety of HOI category settings in different datasets. Therefore, knowledge of the specific HOIs is difficult to transfer. We will compare and evaluate the transferability of interactiveness knowledge and HOI knowledge in Section~\ref{sec:experiment}.

For better representation of the transferability and performance enhancement of interactiveness, we set several transfer learning modes, referred as ``$\mathbf{R}\mathbf{P}_{Tn}\mathbf{C}_{D}$'',
where ``$\emph{T}$'' indicates ``Transfer'', and ``$\emph{n}$'' means $\mathbf{P}$ learns interactiveness knowledge from ``\emph{n}'' datasets:
\noindent{\bf 1) $\mathbf{R}\mathbf{P}_{T1}\mathbf{C}_{D}$}: train $\mathbf{P}$ on 1 dataset and apply $\mathbf{P}$ to another dataset.
\noindent{\bf 2) $\mathbf{R}\mathbf{P}_{T2}\mathbf{C}_{D}$}: train $\mathbf{P}$ on 2 datasets and apply $\mathbf{P}$ to them respectively. 

To compare the transferability of interactiveness knowledge and HOIs knowledge, we set a transfer learning mode ``$\mathbf{R}\mathbf{C}_{T}$'' for $\mathbf{C}$:
\noindent{\bf 3) $\mathbf{R}\mathbf{C}_{T}$:} train $\mathbf{C}$ (without $\mathbf{P}$) on one dataset and apply it to another dataset. For example, we first train and test $\mathbf{C}$ on HICO-DET (referred as ``$\mathbf{R}\mathbf{C}_{D}$''). 
Second, we replace the last FC layer of $\mathbf{C}$ with a FC layer that fits the number of V-COCO HOIs, then finetune $\mathbf{C}$ for 1 epoch on V-COCO train set. 
Last, we test this new $\mathbf{C}$ on V-COCO test set.
Details of the above modes can be found in Table~\ref{tab:mode}.

\subsection{Testing with Non-Interaction Suppression}
\label{sec:test}
After the interactiveness learning,  we further utilize $\mathbf{P}$ to suppress the non-interactive pair candidates in testing, \ie Non-Interaction Suppression (NIS). The inference process is based on tree structure as shown in Figure \ref{Figure:graph-refine}. Detected instances in test set will be paired exhaustively, so a dense graph $\mathcal{G}$ of human and objects is generated. First, we employ $\mathbf{P}$ to compute the interactiveness score of all edges. Next, we suppress the edges that meet NIS conditions, \ie interaction score $s^{\mathbf{P}}_{(h,o)}$ smaller than a certain threshold $\alpha$.

Through NIS, we can convert $\mathcal{G}$ to $\mathcal{G}^{'}$ where $\mathcal{G}^{'}$ denotes the approximate sparse HOI graph. 
The HOI classification score vector $\mathcal{S}^{\mathbf{C}}_{(h, o)}$ of $(v_h, v_o)$ from $\mathbf{C}$ is:
\begin{eqnarray}
    \mathcal{S}^{\mathbf{C}}_{(h,o)} = \mathcal{F}_{{\mathbf{C}}}[\Gamma^{'}; \mathcal{G}^{'}(v_h, v_o)],
\label{eq:C_sparse_score}
\end{eqnarray}
where $\Gamma^{'}$ are input features. The final HOI score vector of a pair $(v_h, v_o)$ can be obtained by:
\begin{eqnarray}
\label{eq:final-score}
    \mathcal{S}_{(h,o)} = \mathcal{S}^{\mathbf{C}}_{(h,o)} \ast s^{\mathbf{P}}_{(h,o)}.
\label{eq:pair_final_score}
\end{eqnarray}
Here we multiply interactiveness score $s^{\mathbf{P}}_{(h,o)}$ from $\mathbf{P}$ by the output of $\mathbf{C}$.

\section{Experiments}
\label{sec:experiment}
In this section, we first introduce datasets and metrics adopted and then give implementation details of our framework. Next, we report our HOI detection results quantitatively and qualitatively compared with state-of-the-art approaches. Finally, we conduct ablation studies to validate the validity of the components in our framework. 

\subsection{Datasets and Metrics}
\noindent{\bf Datasets.} We adopt two HOI datasets HICO-DET~\cite{hicodet} and V-COCO~\cite{vcoco}.
{\bf HICO-DET~\cite{hicodet}} includes 47,776 images (38,118 in train set and 9658 in test set), 600 HOI categories on 80 object categories (same with ~\cite{coco}) and 117 verbs, and provides more than 150k annotated human-object pairs.
{\bf V-COCO~\cite{vcoco}}
provides 10,346 images (2,533 for training, 2,867 for validating and 4,946 for testing) and 16,199 person instances. Each person has annotations for 29 action categories (five of them have no paired object).
The objects are divided into two types: ``object'' and ``instrument''.

\noindent{\bf Metrics.} We follow the settings adopted in~\cite{hicodet}, \ie a prediction is a true positive only when the human and object bounding boxes both have IoUs larger than 0.5 with reference to ground truth, and the HOI classification result is accurate. The role mean average precision~\cite{vcoco} is used to measure the performance.

\subsection{Implementation Details}
We employ a Faster R-CNN~\cite{faster-rcnn} with ResNet-50~\cite{resnet} as $\mathbf{R}$ and keep it frozen.
$\mathbf{C}$ consists of three streams similar to~\cite{hicodet,gao2018ican}, extracting features $\mathbf{\Gamma^{'}}$ from instance appearance, spatial location as well as context. Within human and object streams, a residual block~\cite{resnet} with global average pooling and four 1024 sized FCs are used. Relatively, the spatial stream is composed of two convolutional layers with max pooling, and two 1024 sized FCs. Following~\cite{hicodet,gao2018ican}, we use the late fusion strategy in $\mathbf{C}$. 
$\mathbf{P}$ also consists of three streams (seen in Figure \ref{Figure:overview}). A residual block~\cite{resnet} with global average pooling, and two 1024 sized FCs are adopted in human and object streams.
Residual blocks within these two streams will share weights with those in $\mathbf{C}$.
Spatial-Pose stream consists of two convolutional layers with max pooling and two 1024 sized FCs. The outputs of three streams are concatenated and passed through two 1024 sized FCs to perform interactiveness discrimination.

For a fair comparison, we adopt the object detection results and COCO~\cite{coco} pre-trained weights from~\cite{gao2018ican} which are provided by authors.
Since NIS and LIS can suppress non-interactive pairs, we set detection confidence thresholds lower than~\cite{gao2018ican}, \ie 0.6 for human and 0.4 for object. 
The image-centric training strategy~\cite{faster-rcnn} is also applied. In other words, pair candidates from one image make up the mini-batch. 
We adopt SGD and set an initial learning rate as 1e-4, weight decay as 1e-4, momentum as 0.9. In training, the ratio of positive and negative samples is 1:3. We jointly train the framework for 25 epochs. In LIS mentioned in Equation~\ref{eq:D_sp_func}, we set $T=8.4,k=12.0,w=10.0$.
In testing, the interactiveness threshold $\alpha$ in NIS is set as 0.1. All experiments are conducted on a single Nvidia Titan X GPU.

\subsection{Results and Comparisons}
We compare our method with five state-of-the-art HOI detection methods~\cite{hicodet,Shen2018Scaling,Gkioxari2017Detecting,qi2018learning,gao2018ican} on HICO-DET, and four methods~\cite{vcoco,Gkioxari2017Detecting,qi2018learning,gao2018ican} on V-COCO.
The HOI detection result is evaluated with mean average precision. For HICO-DET, we follow the settings in~\cite{hicodet}: Full (600 HOIs), Rare (138 HOIs), Non-Rare (462 HOIs) in Default and Known Object mode. For V-COCO, we evaluate $AP_{role}$ (24 actions with roles). More details can be found in~\cite{hicodet,vcoco}. 
\begin{table}
\centering
\resizebox{0.48\textwidth}{!}{
\begin{tabular}{l c c c}
\hline
Test Set    & Method   & $\mathbf{P}$-Train Set & $\mathbf{C}$-Train Set   \\
\hline
\hline
\multirow{3}{1.8cm}{HICO-DET}    & $\mathbf{R}\mathbf{P}_{D}\mathbf{C}_{D}$    & HICO-DET & HICO-DET  \\
~       & $\mathbf{R}\mathbf{P}_{T1}\mathbf{C}_{D}$        & V-COCO & HICO-DET  \\
~       & $\mathbf{R}\mathbf{P}_{T2}\mathbf{C}_{D}$        & HICO-DET, V-COCO & HICO-DET  \\
\hline
\multirow{2}{1.8cm}{HICO-DET}     & $\mathbf{R}\mathbf{C}_{D}$         & - & HICO-DET  \\
~       & $\mathbf{R}\mathbf{C}_{T}$                     & - & V-COCO         \\
\hline
\multirow{3}{1.8cm}{V-COCO}  & $\mathbf{R}\mathbf{P}_{D}\mathbf{C}_{D}$  & V-COCO  & V-COCO \\
~       & $\mathbf{R}\mathbf{P}_{T1}\mathbf{C}_{D}$        & HICO-DET & V-COCO   \\
~       & $\mathbf{R}\mathbf{P}_{T2}\mathbf{C}_{D}$        & HICO-DET, V-COCO & V-COCO    \\
\hline
\multirow{2}{1.8cm}{V-COCO}  & $\mathbf{R}\mathbf{C}_{D}$      & -  & V-COCO      \\
~       & $\mathbf{R}\mathbf{C}_{T}$                     & - & HICO-DET     \\
\hline
\end{tabular}}
\caption{Mode settings in experiments.}
\label{tab:mode}
\vspace{-0.3cm}
\end{table}
                          
\noindent{\bf Default Mode.} From Table \ref{tab:hico-det}, we can find that the $\mathbf{R}\mathbf{P}_{D}\mathbf{C}_{D}$ has already outperformed compared methods. We respectively achieve
\textbf{17.03} and \textbf{19.17} mAP 
on Default and Know Object Full sets on HICO-DET. In particular, we boost the performance of
\textbf{2.97} and \textbf{4.18} mAP 
on Rare sets. 
To illustrate, as the generalization ability of interactiveness is beyond HOI category settings, information scarcity and learning difficulty of rare categories is alleviated. So the performance difference between rare and non-rare categories is accordingly reduced.
Results on V-COCO are shown in Table \ref{tab:vcoco}. $\mathbf{R}\mathbf{P}_{D}\mathbf{C}_{D}$ also achieves superior performance and outperforms state-of-the-art method~\cite{gao2018ican} (late and early fusion model), yielding 
\textbf{47.8} mAP, 
which quantitatively validates the efficacy of the interactiveness.
Notably, $\mathbf{R}\mathbf{C}_{D}$ shows limited performance when compared with other models containing $\mathbf{P}$. This reveals the performance enhancement ability of interactiveness network $\mathbf{P}$.

\noindent{\bf Transfer Learning Mode.} By leveraging transferred interactiveness knowledge, $\mathbf{R}\mathbf{P}_{T2}\mathbf{C}_{D}$ 
presents great performance improvement and achieves the most state-of-the-art performance. On HICO-DET, $\mathbf{R}\mathbf{P}_{T2}\mathbf{C}_{D}$ surpasses~\cite{gao2018ican} by 
\textbf{2.38}, \textbf{3.06}, and \textbf{2.17} mAP 
on three Default category sets. Meanwhile, it also outperforms~\cite{gao2018ican} by 
\textbf{4.0} and \textbf{3.4} mAP on V-COCO. 
This indicates the good transferability and effectiveness of interactiveness.
Since HICO-DET train set (38K) is much bigger than V-COCO train set (2.5K), improvement is also larger when transferring is performed from HICO-DET to V-COCO. 
As we can see, mode $\mathbf{R}\mathbf{P}_{T1}\mathbf{C}_{D}$ achieves obvious improvement on V-COCO, but it shows a relatively smaller improvement on HICO-DET when compared with mode $\mathbf{R}\mathbf{P}_{D}\mathbf{C}_{D}$.

We also evaluate the transferability of HOIs knowledge. In comparison with $\mathbf{R}\mathbf{C}_{D}$, $\mathbf{R}\mathbf{C}_{T}$ shows a significant performance decrease of 
\textbf{3.14} and \textbf{4.7} mAP 
on two datasets, as shown in Table~\ref{tab:hico-det} and \ref{tab:vcoco}. It proves that interactiveness is more suitable and easier to transfer than HOIs knowledge.

\begin{table}
\centering
\resizebox{0.48\textwidth}{!}{
\begin{tabular}{l  c  c  c  c  c  c  }
\hline
         & \multicolumn{3}{c}{Default}  &\multicolumn{3}{c}{Known Object} \\
Method         & Full & Rare & Non-Rare  & Full & Rare & Non-Rare \\
\hline
\hline
Shen \etal~\cite{Shen2018Scaling} & 6.46  & 4.24  & 7.12   & -   & -   & -\\
HO-RCNN~\cite{hicodet}       & 7.81  & 5.37     & 8.54     & 10.41   & 8.94   & 10.85\\
InteractNet~\cite{Gkioxari2017Detecting} & 9.94  & 7.16     & 10.77     & -   & -   & -\\
GPNN~\cite{qi2018learning}   & 13.11  & 9.34     & 14.23     & -   & -   & -\\
iCAN~\cite{gao2018ican}     & 14.84  & 10.45     & 16.15     & 16.26   & 11.33   & 17.73\\
\hline
$\mathbf{R}\mathbf{C}_{D}$ & 13.75  & 10.23    & 15.45     & 15.34   & 10.98  & 17.02 \\
$\mathbf{R}\mathbf{P}_{D}\mathbf{C}_{D}$              & \textbf{17.03} & \textbf{13.42} & \textbf{18.11} & \textbf{19.17} & \textbf{15.51} & \textbf{20.26}\\
\hline
\hline
$\mathbf{R}\mathbf{C}_{T}$ & 10.61  & 7.78     & 11.45     & 12.47   & 8.87   & 13.54\\
$\mathbf{R}\mathbf{P}_{T1}\mathbf{C}_{D}$               & \textbf{16.91} & \textbf{13.32} & \textbf{17.99} & \textbf{19.05} & \textbf{15.22} & \textbf{20.19}\\
$\mathbf{R}\mathbf{P}_{T2}\mathbf{C}_{D}$               & \textbf{17.22} & \textbf{13.51} & \textbf{18.32} & \textbf{19.38} & \textbf{15.38} & \textbf{20.57}\\
\hline
\end{tabular}}
\caption{Results comparison on HICO-DET~\cite{hicodet}. $D$ indicates the default mode, and $T$ means the transfer learning model.}
\label{tab:hico-det}
\vspace{-0.5cm}
\end{table}

\noindent{\bf Non-Interaction Reduction.} The non-interactive pairs reduction effect after employing NIS are shown in Table \ref{tab:transfer}.
In default mode $\mathbf{R}\mathbf{P}_{D}\mathbf{C}_{D}$, NIS shows obvious effectiveness. 
With interactiveness transferred from multiple datasets, $\mathbf{R}\mathbf{P}_{T2}\mathbf{C}_{D}$
achieves better suppressive effect and discards 
\textbf{70.94\%} and \textbf{73.62\%} 
non-interactive pairs respectively on two datasets, thus bringing more performance gain.
Meanwhile, $\mathbf{R}\mathbf{P}_{T1}\mathbf{C}_{D}$ also performs well and suppresses a certain amount of non-interactive pair candidates. This suggests the good transferability of interactiveness.
\begin{figure}[!ht]
	\begin{center}
		\includegraphics[width=0.47\textwidth]{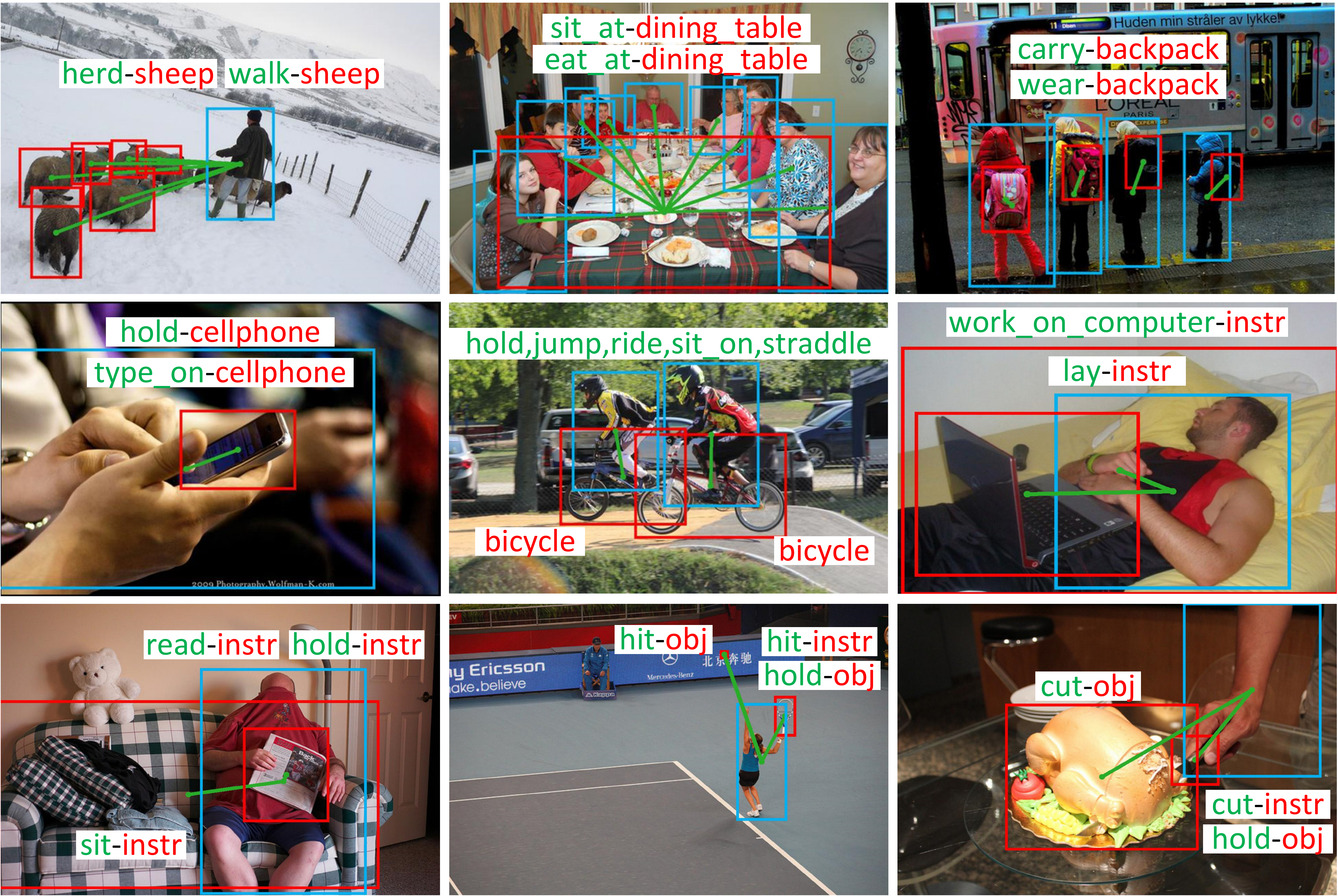}
	\end{center}
	\caption{Visualization of sample HOI detections. Subjects and objects are represented with blue and red bounding boxes. While interactions are marked by green lines linking the box centers.}
	\label{Figure:vis}
	\vspace{-0.3cm}
\end{figure}

\begin{figure*}[!ht]
	\begin{center}
		\includegraphics[width=0.95\textwidth]{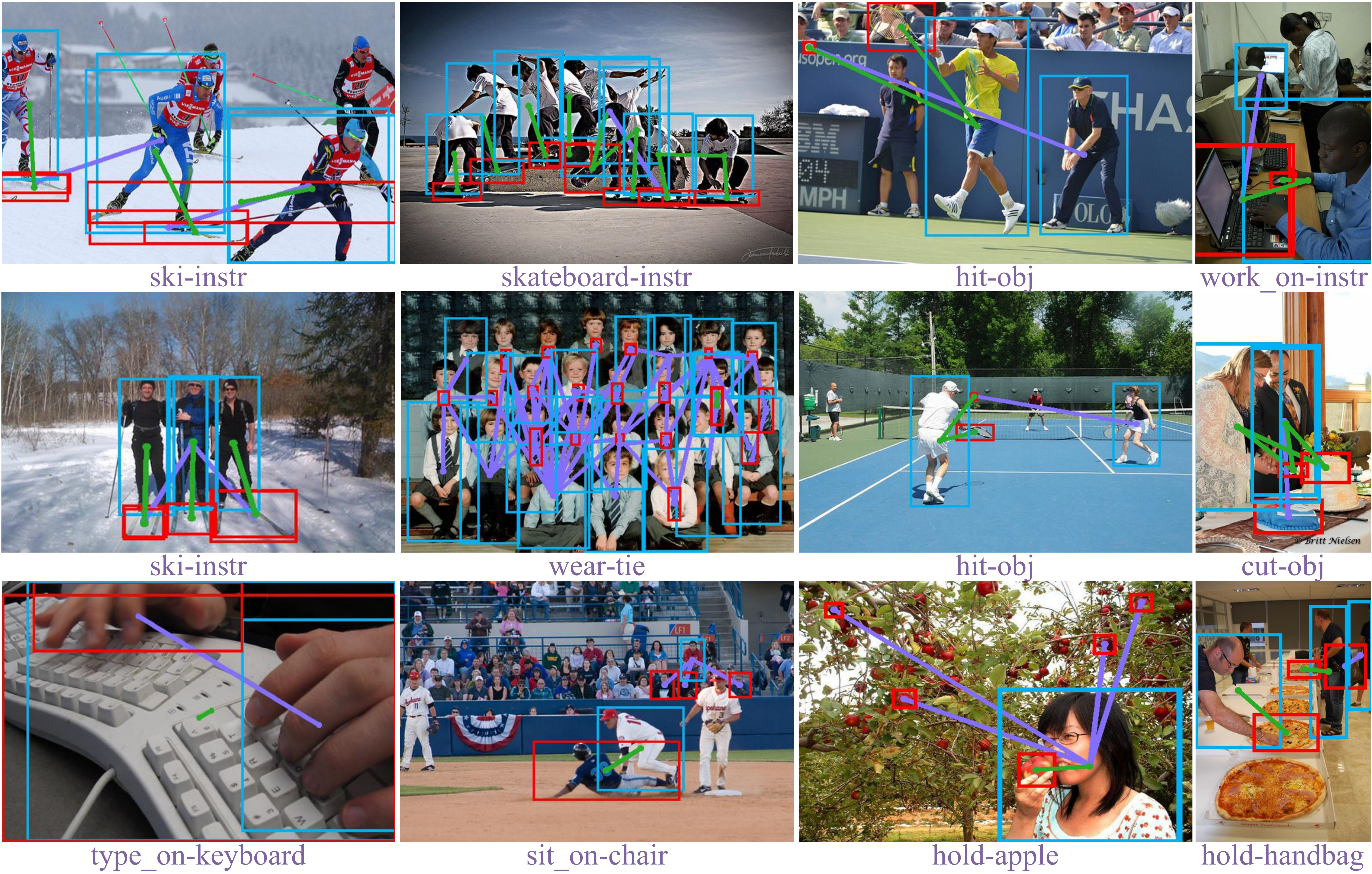}
	\end{center}
	\caption{Visualized effects of NIS. Green lines mean accurate HOIs, while purple lines mean non-interactive pairs which are suppressed. Without NIS, $\mathbf{C}$ would generate false positive predictions for these non-interactive pairs in one-stage inference, which are shown by the purple texts below the images.
	Even some extremely hard scenarios can be discovered and suppressed, such as mis-groupings between person and object close to each other, person and object in clutter scene.}
	\label{Figure:nis_vis}
	\vspace{-0.3cm}
\end{figure*}
\begin{table}
\centering
\resizebox{0.25\textwidth}{!}{
\begin{tabular}{l c }
\hline
Method         &$AP_{role}$\\
\hline
\hline
Gupta \etal~\cite{vcoco}                           & 31.8 \\
InteractNet~\cite{Gkioxari2017Detecting}           & 40.0 \\
GPNN~\cite{qi2018learning}                         & 44.0 \\
iCAN w/ late(early)~\cite{gao2018ican}             & 44.7 (45.3) \\
\hline
$\mathbf{R}\mathbf{C}_{D}$       & 43.2 \\
$\mathbf{R}\mathbf{P}_{D}\mathbf{C}_{D}$                         & \textbf{47.8} \\
\hline
\hline
$\mathbf{R}\mathbf{C}_{T}$       & 38.5 \\
$\mathbf{R}\mathbf{P}_{T1}\mathbf{C}_{D}$                        & \textbf{48.3} \\
$\mathbf{R}\mathbf{P}_{T2}\mathbf{C}_{D}$                        & \textbf{48.7} \\
\hline
\end{tabular}}
\caption{Results comparison on V-COCO~\cite{vcoco}. $D$ indicates the default mode, and $T$ means the transfer learning model.} 
\label{tab:vcoco}
\vspace{-0.4cm}
\end{table}

\noindent{\bf Visualized Results.} Representative predictions are shown in Figure \ref{Figure:vis}. We can find that our model is capable of detecting various kinds of complicated HOIs such as multiple interactions within one pair, one person performing multiple interactions with different objects, one object interacted with multiple persons, multiple persons performing different interactions with multiple objects.

Figure \ref{Figure:nis_vis} shows the visualized effects of NIS. We can see that NIS effectively distinguish the non-interactive pairs and suppress them in extremely difficult scenarios, such as a person performing a confusing action and the tennis ball, a crowd of people with ties. In the bottom-left corner we show an even harder sample. When the subject and object are the left hand and right hand, $\mathbf{C}$ predicts wrong HOI ``type\_on keyboard''. $\mathbf{C}$ may mistake the left hand for the keyboard because they are too close. However, $\mathbf{P}$ accurately figures out that two hands are non-interactive. These results prove that the one-stage method would yield many false positives without interactiveness and NIS.

\subsection{Ablation Studies}
\label{sec:ablation}
In mode $\mathbf{R}\mathbf{P}_{D}\mathbf{C}_{D}$, we analyze the significance of Low-grade Instance Suppressive, Non-Interaction Suppression and the three streams within $\mathbf{P}$ (seen in Table \ref{tab:ablation}).

\noindent{\bf Non-Interaction Suppression} NIS plays a key role to reduce the non-interactive pairs. We evaluate its impact by removing NIS during testing. In other words, we directly use the $\mathcal{S}_{(h,o)}$ from Equation \ref{eq:final-score} as the final predictions without NIS. Consequently, the model shows an obvious performance degradation, which proves the importance of NIS.

\noindent{\bf Low-grade Instance Suppressive} LIS suppress the low-grade object detections and reward the high-grade ones. By removing $L(s_h, s_o)$ in Equation \ref{eq:d_score}, we observe a degradation in Table \ref{tab:ablation}. This suggests that LIS is capable of distinguishing the low-grade detections and improves the performance without using more costly superior object detector. 

\begin{table}
\centering
\resizebox{0.28\textwidth}{!}{
\begin{tabular}{l  c c c}
\hline
Test Set      & Method                  & Reduction\\
\hline
\hline
\multirow{3}{1.8cm}{HICO-DET}  & $\mathbf{R}\mathbf{P}_{D}\mathbf{C}_{D}$      & -65.96\% \\
~   & $\mathbf{R}\mathbf{P}_{T1}\mathbf{C}_{D}$                    & -62.24\%\\
~   & $\mathbf{R}\mathbf{P}_{T2}\mathbf{C}_{D}$                    & -70.94\%\\
\hline
\multirow{3}{1.8cm}{V-COCO} & $\mathbf{R}\mathbf{P}_{D}\mathbf{C}_{D}$      & -65.98\%\\
~  & $\mathbf{R}\mathbf{P}_{T1}\mathbf{C}_{D}$          & -59.51\%\\
~  & $\mathbf{R}\mathbf{P}_{T2}\mathbf{C}_{D}$         & -73.62\%\\
\hline
\end{tabular}}
\caption{Non-interactive pairs reduction after performing NIS.}
\label{tab:transfer}
\vspace{-0.3cm}
\end{table}

\begin{table}
\centering
\resizebox{0.4\textwidth}{!}{
\begin{tabular}{l  c c c}
\hline
         & \multicolumn{2}{c}{HICO-DET}  &V-COCO \\
Method         & Default Full &  \multicolumn{1}{c}{KO Full} & $AP_{role}$ \\
\hline
\hline
$\mathbf{R}\mathbf{P}_{D}\mathbf{C}_{D}$    &17.03      &19.17    &47.8\\
\hline
w/o NIS                 &15.86      &17.35    &46.2\\
w/o LIS                 &16.35      &18.83    &47.4\\
w/o NIS \& LIS          &15.45      &17.31    &45.8\\
\hline
H Stream Only          &14.91      &16.21    &44.5\\
O Stream Only          &15.28      &16.89    &45.2\\
S-P Stream Only        &15.73      &17.46    &46.0\\
\hline
\end{tabular}}
\caption{Results of ablation studies. Human, object, spatial-pose stream are representated as H, O and S-P stream.}
\label{tab:ablation}
\vspace{-0.3cm}
\end{table}

\noindent{\bf NIS \& LIS} Without NIS and LIS both, our method only takes effect in the joint training of $\mathbf{P}$ and $\mathbf{C}$. As we can see in Table \ref{tab:ablation}, performance degrades greatly but still outperforms other methods, which indicates the enhancement brought by $\mathbf{P}$ in the hierarchical joint training.

\noindent{\bf Three Streams.} By keeping one stream in $\mathbf{P}$ each time, we evaluate their contributions as shown in Table~\ref{tab:ablation}.
We can find that spatial-pose stream is the largest contributor, but we still need appearance features from the other two streams to achieve better performance.

\section{Conclusion}
In this paper, we propose a novel method to learn and utilize the implicit interactiveness knowledge,
which is general and beyond HOI categories. Thus, it can be transferred across datasets. 
With interactiveness knowledge, we exploit an interactiveness network to perform Non-interaction Suppression before HOI classification in inference. Extensive experiment results show the efficacy of interactiveness. By combining our method with existing detection models, we achieve state-of-the-art results on HOI detection.

{\small
\paragraph{Acknowledgement:} This work is supported in part by the National Key R\&D Program of China, No.2017YFA0700800, National Natural Science Foundation of China under Grants 61772332. 
}

{\small
\bibliographystyle{ieee}
\bibliography{egbib}
}

\end{document}